\documentclass[runningheads]{llncs}

\usepackage{eccv}

\usepackage{eccvabbrv}

\usepackage{graphicx}
\usepackage{booktabs}
\usepackage{makecell}
\usepackage{xcolor}  
\usepackage{amssymb,mathrsfs,amsmath}
\usepackage{bbding}
\usepackage{ulem}
\usepackage{colortbl}
\usepackage{color}
\usepackage{multirow}
\usepackage{multicol}
\usepackage{wrapfig}

\usepackage[accsupp]{axessibility}  

\usepackage[pagebackref,breaklinks,colorlinks,citecolor=eccvblue]{hyperref}

\usepackage{orcidlink}
\usepackage{geometry}

\begin{document}

\title{\raisebox{-0.2em}{\includegraphics[width=0.6cm]{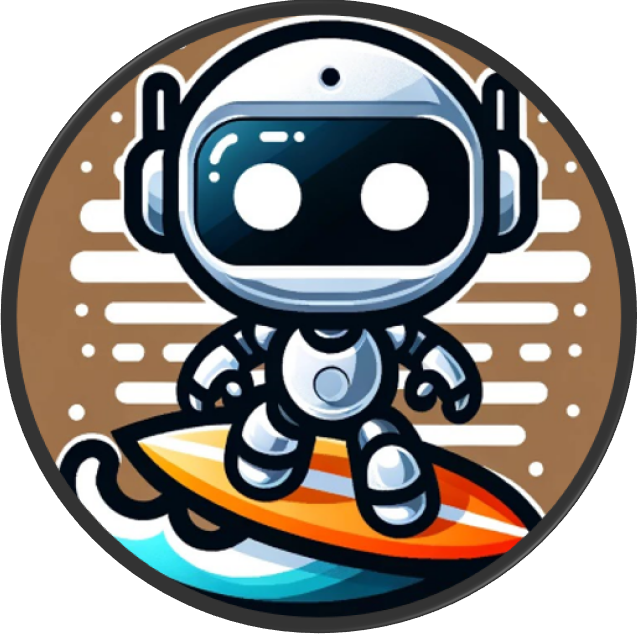}} Surfer: Progressive Reasoning with World Models for Robotic Manipulation} 

\titlerunning{Surfer}


\newcommand*\samethanks[1][\value{footnote}]{\footnotemark[#1]}
\author{Pengzhen Ren\inst{1}\thanks{Equal contribution.} \and
Kaidong Zhang\inst{1}\samethanks \and
Zixuan Li\inst{1} \and
Hetao Zheng\inst{1} \and
Yuhang Wen\inst{1} \and
Fengda Zhu\inst{2} \and
Mas Ma\inst{3} \and
Xiaodan Liang\inst{1,4}\thanks{Corresponding authors.}
}

\authorrunning{P.~Ren et al.}

\institute{Sun Yat-sen University \and
Monash University \and
Dataa Robotics \and
MBZUAI
}

\maketitle

\begin{abstract}
Considering how to make the model accurately understand and follow natural language instructions and perform actions consistent with world knowledge is a key challenge in robot manipulation. This mainly includes human fuzzy instruction reasoning and the following of physical knowledge.
Therefore, the embodied intelligence agent must have the ability to model world knowledge from training data.
However, most existing vision and language robot manipulation methods mainly operate in less realistic simulator and language settings and lack explicit modeling of world knowledge.
To bridge this gap, we introduce a novel and simple robot manipulation framework, called \textbf{Surfer}. 
It is based on the world model, treats robot manipulation as a state transfer of the visual scene, and decouples it into two parts: action and scene. 
Then, the generalization ability of the model on new instructions and new scenes is enhanced by explicit modeling of the action and scene prediction in multi-modal information.
In addition to the framework, we also built a robot manipulation simulator that supports full physics execution based on the MuJoCo physics engine.
It can automatically generate demonstration training data and test data, effectively reducing labor costs.
To conduct a comprehensive and systematic evaluation of the robot manipulation model in terms of language understanding and physical execution, we also created a robotic manipulation benchmark with progressive reasoning tasks, called \textbf{SeaWave}.
It contains 4 levels of progressive reasoning tasks and can provide a standardized testing platform for embedded AI agents in multi-modal environments.
Extensive experiments show that Surfer consistently outperforms all baselines by a significant margin in all manipulation tasks.
On average, Surfer achieved a success rate of 54.74\% on the defined four levels of manipulation tasks, exceeding the best baseline performance of 47.64\%.

  \keywords{Robot Manipulation \and Large Language Model \and World Model}
\end{abstract}


\section{Introduction}
\begin{figure}[t]
    \centering
    \subfloat[RT-1 \textit{vs.} Surfer]{
    \includegraphics[width=0.46 \linewidth]{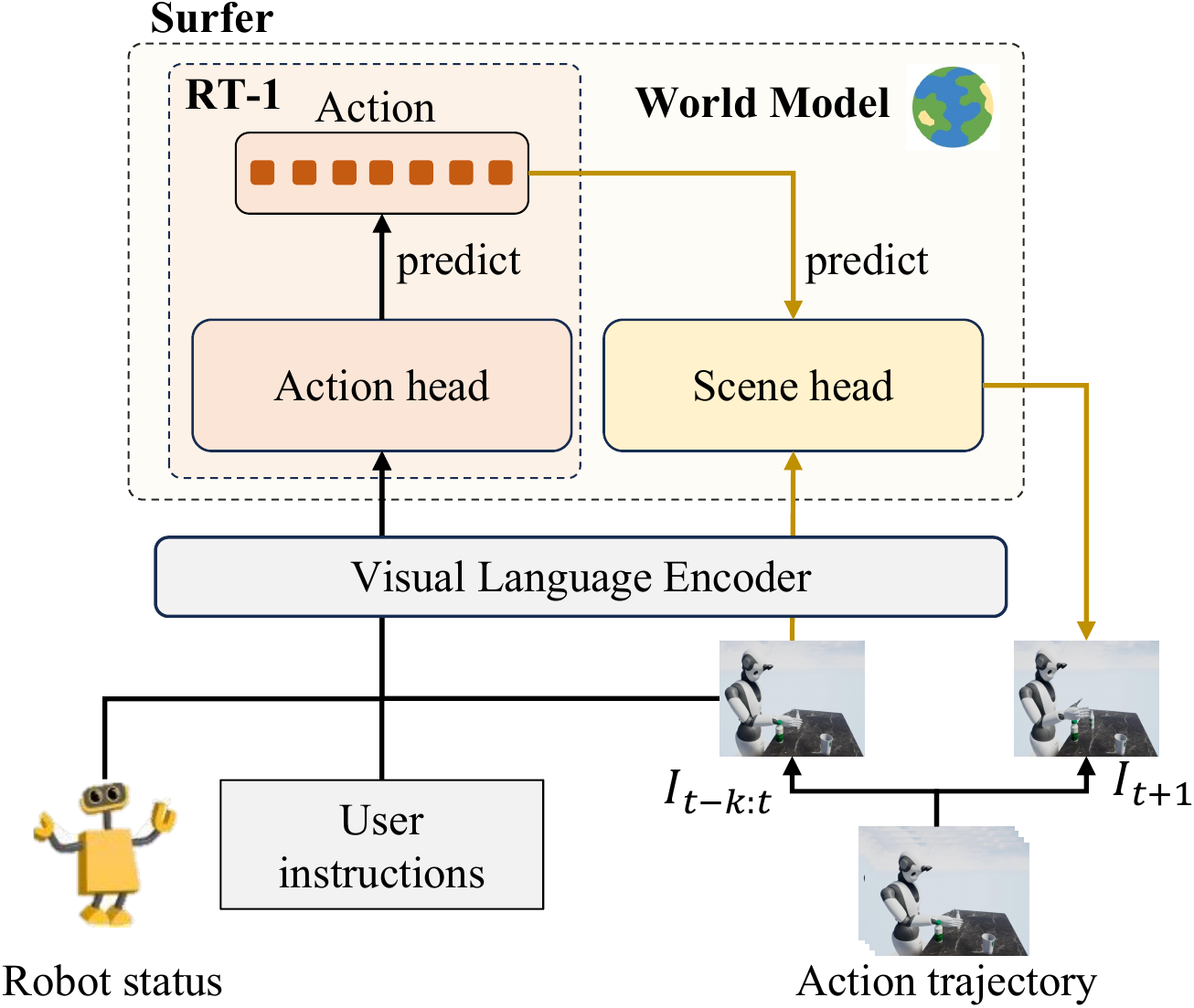}
    \label{fig:surfer_rt1}
    }
    \subfloat[Performance comparison.]{
    \includegraphics[width=0.48 \linewidth]{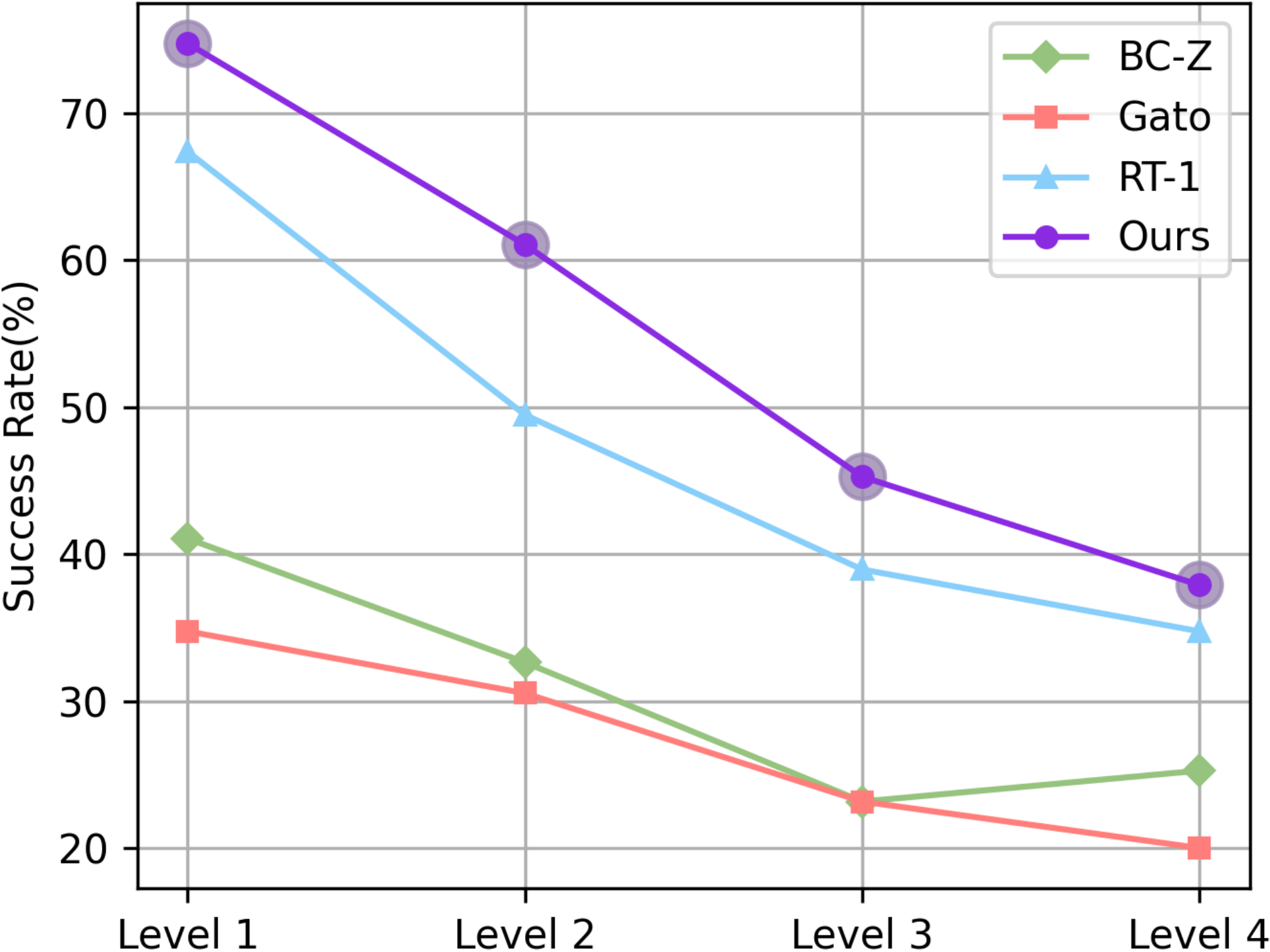}
    \label{fig:result}
    }
    \caption{(a) A brief comparison of RT-1 and Surfer. (b) Comparison of manipulation success rates of different models on four progressive reasoning tasks.}
    
    \vspace{-2em}
\end{figure}

\vspace{-0.5em}
Robot manipulation (RM) is the key to achieving embodied artificial intelligence, which requires robots to understand and follow user instructions, and make reasonable action feedback based on their own state and external vision. It usually requires the robot to have strong generalization capabilities to new instructions and new scenes. The previous end-to-end robot learning \cite{kalashnikov2018scalable, zhang2018deep, kalashnikov2021mt, pmlr-v164-jang22a} was usually designed to perform specific tasks, so the model lacked generalization ability for new tasks.
This requires assembling large dataset and designing the right model. To this end, RT-1 \cite{rt12022arxiv} spent a lot of time collecting a large amount of real machine data and designed a universal robot manipulation model based on the transformer. However, as shown in Figure \ref{fig:surfer_rt1}, it ignores the learning of changes in the robot's manipulation scene, resulting in unsatisfactory robot manipulation performance.
Recently, with the rapid development of multi-modal large models \cite{clipradford2021learning, alignjia2021scaling}, it encourages people to combine various visual-linguistic information with robot manipulation, helping to improve the ability of embodied AI agents to perceive the external environment and parse instructions.
Inspired by this, RT-2 \cite{zitkovich2023rt2}, RT-X \cite{vuong2023openrt-x}, and RoboFlamingo \cite{li2023vision} attempt to combine robot manipulation actions with large-scale visual language data to help the model achieve better generalization ability. 
However, although they used a large amount of multi-modal data, they did not simultaneously model the robot's action execution and manipulation scene changes. 
Similarly, GR-1 \cite{wu2023unleashing} first performs video prediction pre-training on large-scale video data, and then fine-tunes on robot data. 
However, GR-1 \cite{wu2023unleashing} did not explore the logical connection between robot manipulation and scene changes, which is not conducive to the model's understanding of the impact that physical manipulation may have on real-world scene changes, resulting in a lack of predictive ability for the physical world scene changes that may arise from action execution.

In addition, reinforcement-based agents usually require too much interaction with the environment to learn successful manipulation skills \cite{wu2023daydreamer}, which is not allowed in many environments.
The world model \cite{hafner2019dream} allows for planning and behavioral learning with minimal real-world interaction by predicting future outcomes \cite{gal2016improving, ebert2018visual}. It can summarize general dynamic knowledge about the environment and effectively generalize it to a wide range of downstream tasks \cite{sekar2020planning}.
To this end, the latest work attempts to combine world models into tasks such as video games \cite{hafner2020mastering, kim2020learning} and robot control tasks \cite{seo2023masked, kim2020learning, wu2023daydreamer} to reduce the cost of trial and error in the environment and improve learning efficiency and accuracy. 
For example, DayDreamer \cite{wu2023daydreamer} deeply integrates the world model and the robot to help the model achieve faster learning. 
RAP \cite{hao2023reasoning} attempts to redefine LLMs as world models to perform logical reasoning for tasks such as mathematical reasoning, embodied environment plan generation, etc. 
MWM \cite{seo2023masked} builds a world model based on masked autoencoders \cite{he2022maskedMAE} to help robots learn universal visual representations.
However, none of them studied the impact of world models on robot manipulation in complex visual and linguistic environments.

Based on the above observations, in this paper, we propose a novel and simple robot manipulation model with a world model, called \textbf{Surfer}.
We regard the robot's action execution as the state transition of the robot's visual scene and decouple it into two parts: action and scene.
Then, it can explicitly model robot actions and scene changes simultaneously based on robot manipulation data, and directly learn manipulation decisions that conform to world logic knowledge from robot manipulation demonstration videos containing complex user instructions.
This modeling approach to action and scene prediction (AP and SP) can be briefly expressed as:\\
\centerline{\textit{AP}: $\text{historical frame} + \text{user instruction} + \text{robot states} = \text{predicted action}$,}\\ 
\centerline{\textit{SP}: $\text{historical frame} + \text{predicted action} = \text{next frame}$.}\\
Similar to how humans have the ability to foresee changes that their actions may cause to the environment, Surfer can help the model obtain the ability to foresee changes in the scene of action execution through the logical association of AP and SP, thereby improving the model's ability to predict actions.
Specifically, as shown in Figure \ref{fig:surfer_rt1}, the world model built by Surfer mainly includes two modules: (i) \textit{action prediction module} and (ii) \textit{scene prediction module}. 
The former takes user instructions and the current status and visual scene of the robot as input to predict the robot's next action. The latter takes the robot's current visual scene and the output of the action prediction module as input to perform scene prediction.
In this way, Surfer can simultaneously acquire language understanding, action prediction, and visual scene modeling abilities from demonstration videos, thereby improving the model's generalization capabilities in new instructions and new environments.



In order to comprehensively and effectively evaluate the proposed robot manipulation model, this requires that the manipulation benchmark must include a real simulation environment and a large number of complex user instructions as well as sufficient training and testing data.
However, existing benchmarks \cite{NEURIPS2021_021bbc7e,9636667,li2021igibson, gu2023maniskill2} designed for robot manipulation mainly focus on the type of manipulation tasks and the number of object types. It is expected to cover as many manipulation tasks as possible.
There are also some benchmarks designed \cite{ALFRED20, zheng2022vlmbench, NEURIPS2022_4eb33c53} for natural language instruction understanding, but they lack the understanding and reasoning of key visual information (such as appearance, positional relationships, function, etc.), and manipulation tasks are often based on encapsulated action interfaces rather than physics engines, which makes it difficult for the manipulations to directly generalize to the real world.
To address the above challenges, this paper introduces a new fully physically executed robot manipulation benchmark based on the MuJoCo physics engine with progressive reasoning tasks, called \textbf{SeaWave}.
It has multi-level language instruction settings, realistic physical simulation, and the ability to automatically generate demonstration training data and test data.
Specifically, we built a new digital twin scene based on Unreal Engine 5, which includes 8 skills, 139 object categories and four difficulty levels of progressive reasoning manipulation tasks.

Finally, we conduct a comprehensive evaluation of Surfer on progressive reasoning tasks at four difficulty levels. 
As shown in Figure \ref{fig:result}, Surfer significantly outperforms other baselines in all four levels of manipulation tasks.
Specifically, compared to RT-1, Surfer improves the success rate of manipulation by an average of 7.1\% on the four-level tasks defined in the SeaWave benchmark and improves by 3.15\% on level 4 tasks that require close integration of vision and language for reasoning.
Overall, this paper proposes a simple and effective robot manipulation model Surfer with a world model, and also builds a robot manipulation benchmark SeaWave based on the MuJoCo physics engine with progressive reasoning tasks.
Our contributions can be summarized as follows:
\begin{itemize}
    \item \textbf{Surfer}: We propose a novel and simple world model-based robot manipulation method that learns action decisions consistent with world knowledge by explicitly modeling action and scene predictions for robot manipulation.
    \item \textbf{Efficient data generation}: We developed a fully physically executed robot manipulation simulator based on the MuJoCo physics engine, which can automatically generate robot manipulation demonstration data containing complex language instructions, effectively alleviating the problem of insufficient robot manipulation training data.
    \item \textbf{SeaWave}: We propose a robot manipulation benchmark with progressive inference tasks, which can automatically generate training and test data and comprehensively evaluate the model's performance on robot manipulation tasks.
\end{itemize}

\section{Related Work}
\vspace{-0.5em}

\noindent \textbf{The World Model in RM}
The world model \cite{hafner2019dream} has attracted attention due to its ability to predict the future state of the environment from time series, and it has been successfully applied in various fields \cite{kim2020learning, seo2023masked, wu2023daydreamer, ha2018recurrent, hafner2019dream}. It helps the model make inferences consistent with world knowledge based on the external environment, thereby reducing learning costs.
For example, DriveDreamer \cite{wang2023drivedreamer} builds a world model for autonomous driving to help the model generate more reasonable driving strategies. DayDreamer \cite{wu2023daydreamer} also attempts to deeply integrate robot learning with the world model to help the model free itself from a large amount of trial and error and improve learning efficiency.
Different from previous work, this paper proposes to treat robot manipulation as a state transfer of the visual scene, decouple it into two parts: actions and scenes, and then explicitly model them simultaneously.
It can help robots make action decisions that are more consistent with world knowledge.

\noindent \textbf{Large Language Models}
Recently, artificial general intelligence (AGI) has made remarkable progress, primarily driven by the emergence of LLMs \cite{alayrac2022flamingo, chowdhery2022palm,openai2023gpt4, touvron2023llama, koala_blogpost_2023, alpaca, du2022glm}. 
Currently, LLMs have demonstrated language processing capabilities that are comparable to or even exceed the average human ability in many fields. The success of LLMs across a broad range of language tasks is largely attributable to rapid advances in architecture design and training methods \cite{openai2023gpt4}.
In addition, researchers have also put considerable effort into the design of LLM's hierarchical prompting system, aimed at eliciting coherent and logical responses \cite {wei2023chainofthought,wang2023selfconsistency,yao2023tree}.
With the rapid iteration of LLMs research, recent studies have explored the potential of harnessing LLMs as plug-ins or knowledge bases to adapt to diverse tasks \cite{wu2023visual,maaz2023videochatgpt}. 
It is a very challenging task to design a reasonable benchmark for a comprehensive and systematic evaluation of the embodied AI agent manipulation model.
This encourages us to use LLMs to help us build a multi-level instruction library for progressive reasoning tasks based on multi-modal inputs.

\noindent \textbf{Robotic Manipulation Benchmark}
Robotic manipulation is an area of great research and application, and many previous studies have proposed corresponding benchmarks \cite{NEURIPS2021_021bbc7e,9636667,li2021igibson,pmlr-v164-srivastava22a}. 
Among them, cross-modal robot manipulation benchmarks \cite{9001253, ALFRED20, ALFWorld20, jiang2023vima, NEURIPS2022_4eb33c53, zheng2022vlmbench, yenamandra2023homerobot} that incorporate target descriptions or human instructions have attracted much attention.
In particular, VLMBench \cite{zheng2022vlmbench} presents a simulator benchmark for object manipulation guided by human language, while HandMeThat \cite{NEURIPS2022_4eb33c53}, Teach \cite{padmakumar2021teach}, DialFRED \cite{gao2022dialfred} further challenge the agents in understanding instructions by introducing ambiguous utterances instead of clear commands.
However, the above benchmarks either lack full physics execution supported by the corresponding physics engines, cannot automatically generate training and test data, or cannot evaluate progressive inference tasks.
To this end, in this paper, we propose a benchmark that supports full physics execution based on the MuJoCo physics engine with a progressive reasoning task setting. 
It can also automatically generate both demonstration training data and test data, which greatly improves the efficiency of robot manipulation data generation.

\begin{figure*}[t]
    \centering
    \includegraphics[width=1 \linewidth]{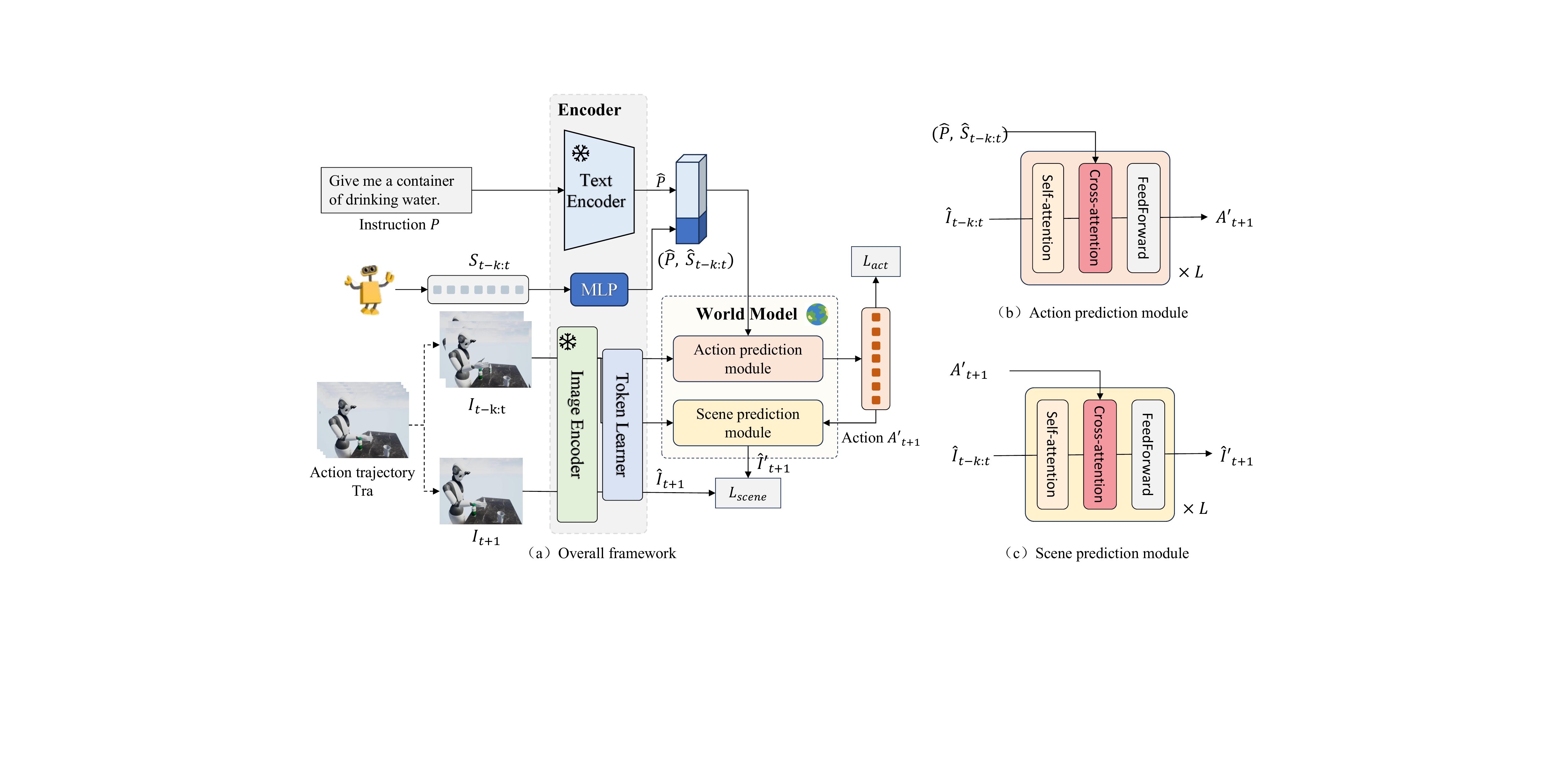}
    \caption{Overall framework of Surfer. It mainly contains two modules: action prediction and scene prediction.}
    \label{fig:model}
    \vspace{-2em}
\end{figure*}

\section{Surfer}
\vspace{-0.5em}
Our goal is to build an embodied AI agent model that can perform manipulations consistent with world logical knowledge under complex visual and linguistic instructions.
To this end, we propose Surfer, a novel robot manipulation with world model reasoning. The overall framework of Surfer is shown in Figure \ref{fig:model}. 

Overall, as shown in Figure \ref{fig:model}a, formally, for any given instruction $P$, we have a set of corresponding robot action trajectories $\text{Tra}=I_{t=1}^{N}$ and states $\mathcal{S}=S_{t=1}^{N}$ and execution actions $\mathcal{A}=A_{t=1}^{N-1}$. Where $N$ represents the length of the action trajectory, and $I$ is the RGB image of the trajectory $\text{Tra}$. Among them, the robot state $\mathcal{S}$ includes seven dimensions for arm movement $(x, y, z, roll, pitch, yaw, gripper)$.
The action $\mathcal{A}$ consists of changes in arm movement and opening of the gripper. And, we have an action frame transformation relationship: 
\begin{equation}
    I_{t-k:t} \xrightarrow{A_{t+1},P} I_{t+1}.
\end{equation}
For step $t$, we have $I_{t-k:t}$, $S_{t-k:t}$, and corresponding instruction $P$, and encode them through their respective encoders to get image feature $\hat{I}_{t-k:t}$, robot state feature $\hat{S}_{t-k:t}$, and text feature $\hat{P}$. $k$ means the length of history.
Finally, the world model takes these features as inputs to make predictions and inferences that match the world's logical knowledge for the robot's actions and the next action frame.

\noindent \textbf{Encoder} As shown in Figure \ref{fig:model}a, our encoder mainly contains three types of inputs: image $\{I_{t-k:t},I_{t+1}\}$, robot state $S_{t-k:t}$, and text $P$.
For RGB image $I$, we use pre-trained CLIP \cite{DBLP:journals/corr/abs-2103-00020} visual encoder for encoding. 
To speed up inference, we use TokenLearner \cite{ryoo2021tokenlearner} to compress the number of tokens.
Specifically, TokenLearner compresses 49 visual tokens output from the CLIP visual encoder into 8 final tokens.
For instructions $P$ and robot state $S_{t-k:t}$, we encode them using a pre-trained CLIP text encoder and a simple MLP, respectively.

\noindent \textbf{World Model} Our world model mainly consists of two modules: action prediction $E_{ap}$ and scene prediction $E_{sp}$. 
As shown in Figure \ref{fig:model}b, the action prediction module is stacked by $L$ standard transformer decoder layers \cite{vaswani2017attention}.
Each standard transformer decoder layer consists of three modules to process query features in the following order: a self-attention module, a cross-attention, and a feed-forward network (FFN).
The action prediction module takes $\hat{I}_{t-k:t}$ as input, $\hat{S}_{t-k:t}$, and $\hat{P}$ are conditioned by a series of cross-attention layers. 
That is, following \cite{DBLP:journals/corr/abs-1910-10683}, we use the encoding $\hat{I}_{t-k:t}$ of the action frames as the input $q$ of the cross-attention layer, and the concat embedding of the instruction encoding $\hat{P}$ and robot state $\hat{S}_{t-k:t}$ as $k$, $v$ to perform action prediction.
The above process can be simply expressed as:
\begin{equation}
    A'_{t+1} = E_{ap}(\hat{I}_{t-k:t}, (\hat{P}, \hat{S}_{t-k:t})).
\end{equation}

In addition, as shown in Figure \ref{fig:model}c, based on the reasoning advantage of the world model on time series, similar to the action prediction module, we use the action frame encoding $\hat{S}_{t-k:t}$ and predicted actions $A_{t-k:t}$ as the input of the scene prediction module to predict the next action frame $\hat{I}'_{t+1}$ and form supervision with the ground truth $\hat{I}_{t+1}$ in the action trajectory. The above process can be simply expressed as:
\begin{equation}
    \hat{I}'_{t+1} = E_{sp}(\hat{I}_{t-k:t}, A'_{t+1})
\end{equation}

\noindent \textbf{Loss.} We optimize model by action prediction loss $L_{act}$ and scene prediction loss $L_{scene}$. We use MSE Loss to compute $L_{act}$. We follow the design of I-JEPA \cite{assran2023selfsupervised}, input the next action frame $I_{t+1}$ to get real feature $\hat{I}_{t+1}$ through CLIP visual encoder and TokenLearner same as used by the current image and then compute the loss $L_{scene}$ using the average $L_2$ distance between the predicted feature $\hat{I}'_{t+1}$ and real feature $\hat{I}_{t+1}$.

\vspace{-1em}
\section{SeaWave}
\vspace{-0.5em}
In this section, we described in detail the basic components of the simulator in the SeaWave benchmark and formalized our environment and tasks using notations and concepts. Finally, we proposed a progressive reasoning task with natural language instructions for robotic manipulation. 
We compare SeaWave to other robotic manipulation benchmarks in detail. As shown in Table \ref{tab:BenchmarkComparison}, SeaWave is the only benchmark that has progressive inference tasks and supports full physical execution and automated data generation.

\begin{figure*}[t]
    \centering
    \includegraphics[width=1 \linewidth]{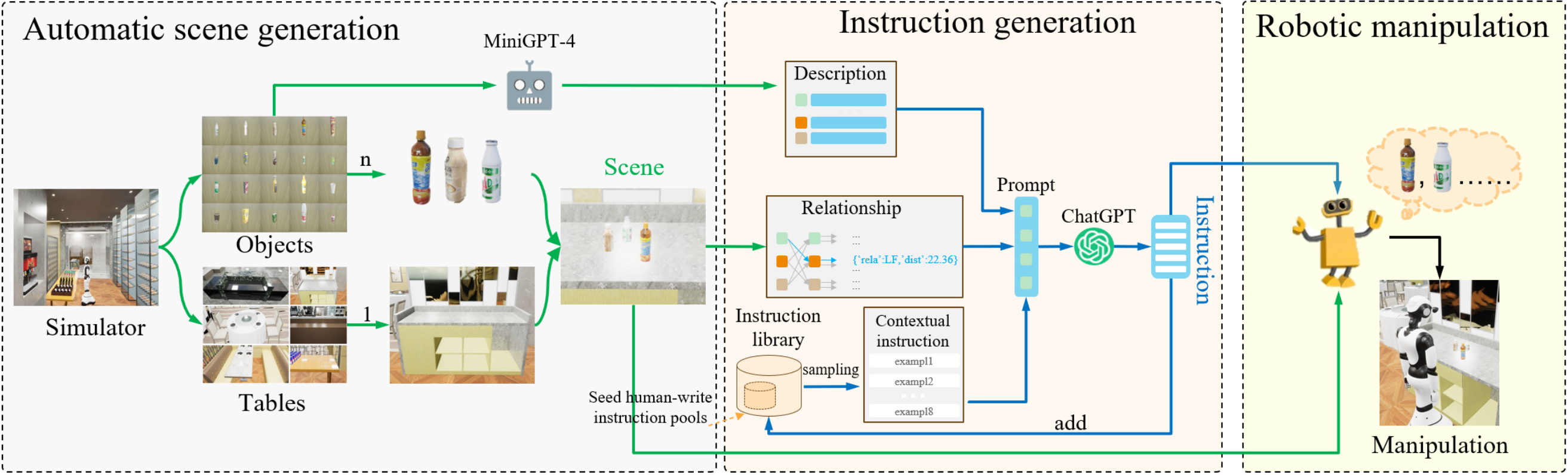}
    \caption{In the SeaWave benchmark, the proposed general pipeline mainly consists of three parts: automatic scene generation, instruction generation, and robotic manipulation.}
    \label{fig:pipeline}
    \vspace{-1.0em}
\end{figure*}

\subsection{Simulator}

\begin{table*}[t]
\centering
\footnotesize
\caption{The list of skills on SeaWave and their description and success condition.}
\resizebox{\columnwidth}{!}{
\begin{tabular}{l|c|c}
\toprule
\textbf{Skill}  & \textbf{Description} & \textbf{Success Condition} \\ \toprule
Pick Target  & Grasp the target object and lift it & The target object is 10cm away from the table \\ 
Place Target  & place the target object on the table & The target object stands upright on table \\
Move A Near B  & Grasp A and move it closer to B  & A is moved and ends up being less than 10cm away from B \\
Open Door  & Open the door & The door is opened more than 80 degrees \\ 
Close Door  & Close the door & The door is closed to less than 10 degrees \\
Push Target Front  & Push the target object forward & The target object is pushed forward 10cm \\ 
Push Target Aside  & Push the target object to the left or right & The target object moves 10cm to the left or right \\ 
Knock Target Over  & Knock the target object over & The target object falls on the table \\ \bottomrule

\end{tabular}
}
\label{tab:skills}
\end{table*}

\begin{figure}
    \centering
    \subfloat[Simulators and examples of progressive reasoning tasks.]{
    \includegraphics[width=0.5 \linewidth]{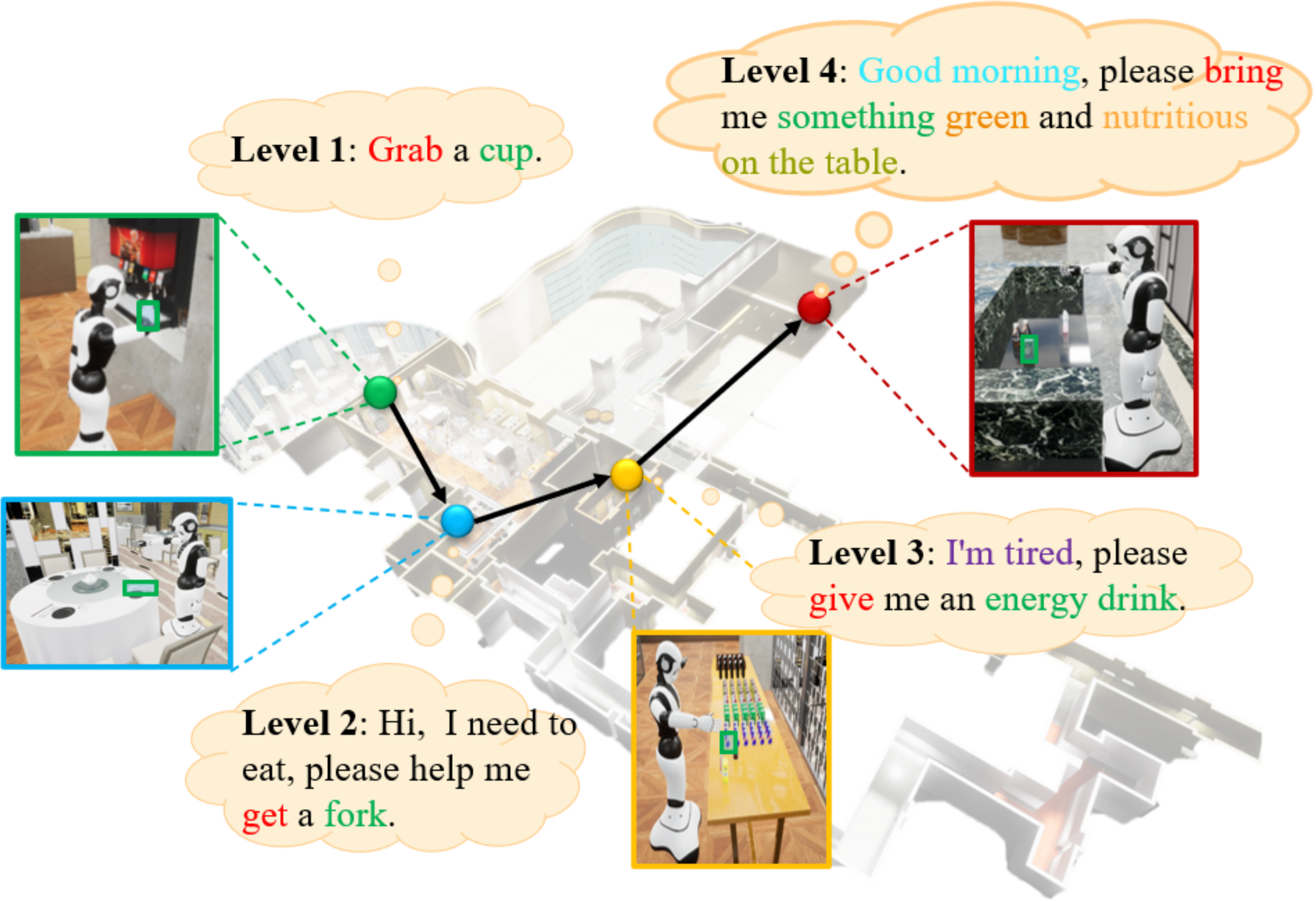}
    \label{fig:multi_level_tasks}
    }
    \hspace{1em}
    \subfloat[A subset of the object library.]{
    \includegraphics[width=0.4 \linewidth]{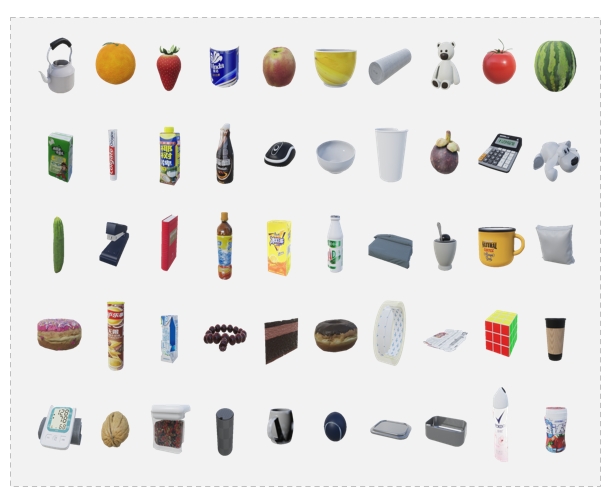}
    }
    \caption{Overview of SeaWave benchmark. (a) The green box indicates the target object that the current instruction requires the robot to grasp. There is only a single object in the level 1 scene, and multiple objects in the level 2, 3, and 4 scenes. The difficulty of the four levels of tasks increases in sequence. In particular, level 4 requires a deep integration of vision and language information to make accurate decisions. 
    (b) SeaWave’s object library contains the most common objects and supports a variety of robot manipulation scenarios.}
    \vspace{-1em}
    \label{fig:simulator}
\end{figure}

To conveniently and fairly compare the success rate of robots understanding and executing human natural language instructions under different baseline models, we built a restaurant scene simulator based on UE5 for the study of robot manipulation.
Figure \ref{fig:simulator} (a) shows the overall scene graph of this restaurant simulator.
It contains 139 object categories, Figure \ref{fig:simulator} (b) shows a subset of our simulator object library, which contains common objects of various shapes and purposes. To add more high-quality articulated objects, the simulator also supports the introduction of PartNetMobility\cite{xiang2020sapien}, currently the highest quality and largest articulated object dataset.
In addition, we define several common robotic manipulation skills to better simulate how robots work in daily environments. These tasks are classified and listed in Table \ref{tab:skills} in detail.
This guarantees the construction of a generic robot manipulation environment and the design of complex and diverse human natural language instructions. 

\begin{table*}[t]
    \centering
    
    \footnotesize
    \caption{Comparison with diverse robotic manipulation benchmarks. "-" indicates that the original paper does not provide the specified total number of object entities (including non-grabbable objects), object categories, or language instructions. "Full-physics" means everything is implemented based on the physics engine, without relying on existing interfaces. “AutoData” means the automatic generation of test data and demonstration training data.}
    \resizebox{\columnwidth}{!}{
    \begin{tabular}{lccccccc}
        \hline 
        Benchmark & Source & \thead{\#Entities\\(\#Category)} & \thead{\#Language\\Instruction} & \thead{Physics\\Engine} & Full-physics & AutoData & \thead{Progressive\\Reasoning}\\
        \hline
        iGibson 1.0 \cite{9636667} & IROS'21 & 570 (-) & 0 & Pybullet & \XSolidBrush & \XSolidBrush  & \XSolidBrush \\
        iGibson 2.0 \cite{li2021igibson} & PMLR'22 & 1217 (-) & 0 & Pybullet & \XSolidBrush & \XSolidBrush & \XSolidBrush \\
        BEHAVIOR \cite{pmlr-v164-srivastava22a} & PMLR'22 & 1217 (391) & 0 & Pybullet & \XSolidBrush & \XSolidBrush & \XSolidBrush \\
        HAB\cite{NEURIPS2021_021bbc7e} & NeurIPS'21 & 32 (2) & 0 & Bullet & \XSolidBrush & \XSolidBrush & \XSolidBrush \\
        Ravens \cite{zeng2020transporter} & CoRL'20 & - (-) & 0 & PyBullet & \XSolidBrush &  \XSolidBrush & \XSolidBrush \\
        Robosuite \cite{robosuite2020} & arXiv'20 & - (-) & 0 & MuJoCo & \Checkmark & \XSolidBrush & \XSolidBrush \\
        HandoverSim \cite{9812302} & ICRA'22 & 20 (-) & 0 & PyBullet & \Checkmark & \XSolidBrush  & \XSolidBrush\\
        ManiSkill \cite{mu2021maniskill} & NeurIPS'21 & 162 (3) & 0 & PhysX & \Checkmark & \XSolidBrush  & \XSolidBrush\\
        ManiSkill2 \cite{gu2023maniskill2} & ICLR'23 & >2144 (-) & 0 & PhysX & \Checkmark & \XSolidBrush  & \XSolidBrush\\
        RLBench \cite{9001253} & RAL'20 & - (-) & - & Bullet/ODE & \Checkmark & \XSolidBrush  & \XSolidBrush\\
        ALFRED \cite{ALFRED20} & CVPR'20 & - (84) & 25K & Unity & \XSolidBrush & \XSolidBrush  & \XSolidBrush\\
        VIMA-BENCH \cite{jiang2023vima} & ICML'23 & 29 (-) & - & PyBullet & \XSolidBrush &  \Checkmark  & \XSolidBrush\\
        VLMBench \cite{zheng2022vlmbench} & NeurIPS'22 & - (-) & - & Bullet/ODE & \XSolidBrush & \Checkmark  & \XSolidBrush\\
        HandMeThat \cite{NEURIPS2022_4eb33c53} & NeurIPS'22 & 200 (-) & - & Unity & \XSolidBrush & \XSolidBrush & \Checkmark\\
        \hline
        SeaWave (Ours) & & 167 (139) & 3.4K & Mujoco& \Checkmark& \Checkmark & \Checkmark\\
        \hline
    \end{tabular}
    }
    
    \vspace{-1em}
    \label{tab:BenchmarkComparison}
\end{table*}

\begin{table*}[t]
    \centering
      
    \footnotesize
    \caption{The setting of progressive reasoning tasks.}
    \resizebox{\linewidth}{!}{
    \begin{tabular}{cccclc}
          \hline 
        Level  & Instruction type & \makecell{Single/Multiple\\Objects} & \makecell{\#Language\\Instruction}  & Capability Assessment & Difficulty\\
        \hline
        1 & \textit{v.}+\textit{n.} & Single & 80  & Basic manipulation & \FiveStar \\
        2 & Natural language & Multiple & 240  & Natural language understanding & \FiveStar \FiveStar\\
        3 & Natural language & Multiple & 858 & Intention inference & \FiveStar\FiveStar\FiveStar \\
        4 & Natural language  & Multiple& 2267  & Visual perception and decision-making & \FiveStar\FiveStar\FiveStar\FiveStar\\
        \hline
    \end{tabular}
    }
    
    \vspace{-1.5em}
    \label{tab:tasks}
\end{table*}

\vspace{-1.5em}
\subsection{Pipeline}
\label{sec: pipeline}

In general, we also designed a general pipeline for the SeaWave benchmark, as shown in Figure \ref{fig:pipeline}, which mainly consists of three parts: automatic scene generation, automatic instruction generation, and robotic manipulation.
\textit{(i)} Automatic scene generation is responsible for generating abundant and diverse scenes for agents to train and test.
\textit{(ii)} Furthermore, the acquisition of large-scale complex and high-quality human natural language instructions is costly and difficult. To this end, we also designed a module that mimics human natural language instruction generation using LLMs (\textit{e.g.}, ChatGPT \cite{ChatGPT}) to provide sufficient and high-quality natural language instructions for robot manipulation.
\textit{(iii)} Finally, the robot is controlled to perform corresponding manipulations by inputting the generated visual scene and natural language instructions to the robot.
In addition, we also provide detailed instructions for automatic robot demonstration generation, which can help us collect training data more conveniently.
Next, we introduce these components in detail.

\noindent \textbf{Automatic Scene Generation} As shown in Figure \ref{fig:pipeline} (left), randomly select a table $T_i \in T=\{T_i\}_i^M$, and randomly select \(n\) objects $\{obj_i(name_i, x_i,y_i,yaw_i,h_i)\}_i^n$ from objects library $O=\{obj_i\}_i^N$, and then randomly put these objects on the table, where $yaw$ represents the orientation of the object. The scene can be represented as  $S=\{\{obj_i(name_i, x_i, y_i, yaw_i, h_i)\}_i^n, T_i\}$, where \(x\) and \(y\) are the coordinates of the object with respect to the robot \((x_0,y_0)\), that is, the actual coordinates of the object $obj_i$ are  \(( x_0+x_i, y_0 + y_i, h_i)\).

\noindent \textbf{Automatic Instruction Generation} To collect sufficient high-quality complex multi-level human-like natural language instructions, we use ChatGPT \cite{ChatGPT} as the instruction generation model. 
Specifically, as shown in Figure \ref{fig:pipeline} (middle), we automatically generated human-like natural language instructions that meet the requirements through the following four steps:\\
\textit{Step 1}: Description generation. Use MiniGPT-4 \cite{zhu2023minigpt} to generate appearance and function descriptions for each object in the object library.\\
\textit{Step 2}: Position relationship calculation. According to the coordinates of each two objects in the scene, calculate the positional relationship between them, \textit{e.g.}, “ADMilk” $\rightarrow$ “IceTea”: $\{rela:$ LF$, dist: 22.36\}$, where LF means LeftFront. Similarly, LB, RF, and RB represent LeftBack, RightFront, and RightBack respectively.\\
\textit{Step 3}: 
Instruction generation.
Following \cite{wang2022self, weller2020learning, mishra2021cross, wei2021finetuned}, 
we generate new instructions from a small set of seed human-written instruction pools in a bootstrapping fashion.
Specifically, in each step, we sample eight instructions from the seed instruction pools as contextual instruction examples (six of them are human-written instructions, and the other two are instructions generated before this step).
Then, these contextual instructions, object descriptions, and positional relationships of the scene are used as inputs to ChatGPT \cite{ChatGPT}, and ChatGPT is required to generate corresponding instructions according to the customized prompts. 
Among them, the prompt setting details are shown in Section 2 of the Appendix.\\
\textit{Step 4}: Instructions update. Finally, following \cite{wang2022self}, the generated instructions are added to the instruction library to increase the diversity of the instruction library.

\noindent \textbf{Robotic Manipulation}
As shown in Figure \ref{fig:pipeline} (right), The robot accepts visual scenes and natural language instructions and follows the instructions to perform corresponding manipulations. 


\noindent \textbf{Automatic Demonstration Generation}
\label{sec:Motion_planner}
To help us better control the robot to perform tasks, we added the inverse kinematics plug-in to the simulator and implemented a motion planner with rapidly-exploring random trees\cite{LaValle1998RapidlyexploringRT}. It can help us effectively control robots for experiments and data collection.
To training by imitation learning, we perform data collection. We selected 14 objects and 10 tables and used the pipeline shown in Figure \ref{fig:pipeline} to generate data. After automated scene generation and instruction generation, we call the motion planner to generate trajectories, judge the generation results, remove failed data, and add successful trajectories to our training set. We collected a total of 13K trajectories. 

\vspace{-1em}
\subsection{Progressive Reasoning Tasks}
\label{sec:PRT_tasks}

Natural language is one of the most direct and effective ways of human-computer interaction.
However, due to the complexity and variability of external visual scenes and human natural language instructions, understanding and executing these instructions has become one of the key challenges in embodied AI research.
To systematically analyze and study these challenges, we classified tasks into four levels according to the complexity of instructions and the ease of operation.

\noindent The specific content is as follows:
\begin{itemize}
    \item Level 1: The scene contains only one object, and the robot receives explicit machine language commands consisting of \textit{verbs + nouns}. It is used to evaluate the basic manipulation capabilities of the model.
    \item Level 2: This task scenario contains multiple objects and the natural language instructions explicitly include the name of the target object. It is used to evaluate the model's ability to understand conventional natural language instructions.
    \item Level 3: This task scene contains multiple objects, but the natural language instructions do not contain the name of the target object, but only provide expressions related to the functionality of the target object. It is used to evaluate the model's ability to infer the intent of human instructions.
    
    \item Level 4: This task scene contains multiple objects. The natural language instructions do not include the name of the target object but only provide expressions related to the function, appearance, or location of the target object. This instruction requires the model to have strong visual and language information processing capabilities at the same time. It is used to evaluate the model's visual perception and decision-making capabilities.
\end{itemize}
Overall, we generated 80, 240, 858, and 2267 instructions for four levels of tasks, respectively. 
We have summarized the main settings for these four levels of tasks in Table \ref{tab:tasks}, and shown corresponding examples in Figure \ref{fig:multi_level_tasks}.

\vspace{-1em}
\section{Experiments}
\vspace{-0.5em}
Our experiments aim to address three key inquiries: 

    (1) How effectively can Surfer learn from the dataset and attain strong generalization capabilities? 
    
    (2) Can Surfer be robust in unseen scenarios?
    
    (3) How do different designs influence the performance of the results?
    
In this section, we first introduce the experimental setup of the SeaWave benchmark in Section \ref{sec:exp_setting}. 
Then, the baseline models used were introduced in Section \ref{sec:baselines}.
Next, in Section \ref{sec:results_rm}, we conducted a detailed evaluation of the robot manipulation tasks defined on the SeaWave benchmark for each baseline. 
We also evaluated the robustness of the model in section \ref{sec:generalization_results}.
Finally, we ablated the main modules of Surfer in section \ref{sec:ablation studies}.



\vspace{-1.5em} 
\subsection{Settings} 
\vspace{-0.5em}
\label{sec:exp_setting}

During training and testing, for each task, we generate corresponding scenes including 1 table ($M=10$) and $n$ (\textit{i.e.}, $1\leq n\leq 3, N=14$) objects for the selected instructions.
Note that $n=1$ in level 1 and $1 < n\leq 3$ other level tasks.
To fairly verify the generalization ability of Surfer, except for the command composed of \textit{verbs} + \textit{nouns} used in level 1, the rest of the instructions used in testing have not been used in training.
Specifically, we selected 80, 160, 686, and 2041 instructions from levels 1, 2, 3, and 4 respectively for training, and the rest for testing.
Finally, the success rate of robot manipulation is used as the evaluation metric for each task.

\vspace{-1.5em}
\subsection{Baselines}
\vspace{-0.5em}
\label{sec:baselines}

In the experiment, we selected the state-of-the-art architectures BC-Z \cite{pmlr-v164-jang22a}, Gato \cite{reed2022generalist}, and RT-1 \cite{rt12022arxiv} as the baseline models for the SeaWave benchmark.
These models can effectively control the robot to complete the specified manipulation tasks by receiving multi-modal instruction information.
We train and test these baseline models in our setting for a comprehensive evaluation of our proposed benchmarks. 
Specifically, the implementation details of the above baseline models are as follows:

\noindent \textbf{BC-Z} \cite{pmlr-v164-jang22a} includes a pre-trained multilingual sentence encoder, a FiLM encoder, and a two-layer MLP to decode robot actions. In experiments, the text encoder is implemented using T5Adapter \cite{2020t5}. 

\noindent \textbf{Gato} \cite{reed2022generalist} defines three distinct approaches to embed images, continuous value sequences, as well as discrete value sequences. In our experiment, images are divided into patches in raster order as Gato does. Later, the patches are embedded using a residual encoder with the predefined patch positional information. No continuous value is provided. Additionally, instructions are given as discrete value sequences, but for fair comparison, T5Adapter \cite{2020t5} is directly adopted to obtain language embeddings, instead of SentencePiece encoding. A decoder-only Transformer serves as the backbone. But in our experiments, as RT-1 \cite{rt12022arxiv} suggests, the model size is limited to be of similar size to BC-Z and RT-1, which means a tiny version of Gato would be used in comparisons.

\noindent \textbf{RT-1} \cite{rt12022arxiv} consists of a Universal Sentence Encoder for instruction embedding, a language-conditioned FiLM \cite{perez2018film} to encode images, a TokenLearner \cite{ryoo2021tokenlearner} to reduce token numbers, and a decoder-only Trasformer to output tokenized robot actions. For a fair comparison, we use the same text encoder adopted in the above baselines.

\vspace{-1.5em}
\subsection{Results on Robotic Manipulation}
\vspace{-0.5em}
\label{sec:results_rm}

\begin{table*}[t]
    \centering
    \caption{Performance comparison of different methods in four levels of manipulation tasks (\%).}
    \vspace{-0.5em}
    \begin{tabular}{l|cccc|c}
    \toprule
        Model &  Level 1 &  Level 2 & Level 3 & Level 4 & Mean \\ \hline
        BC-Z \cite{pmlr-v164-jang22a} & 41.05 & 32.63 & 23.16  & 25.26 & 30.53\\
        Gato \cite{reed2022generalist} & 34.74 & 30.53 & 23.16 & 20.00 & 27.11 \\
        RT-1 \cite{rt12022arxiv} & 67.38 & 49.47 & 38.95  & 34.74 & 47
        .64\\ \hline
        Surfer & \textbf{74.74} $(7.36\uparrow)$  & \textbf{61.05} $(11.58\uparrow)$ & \textbf{45.26} $(6.31\uparrow)$ & \textbf{37.89} $(3.15\uparrow)$ & \textbf{54.74} $(7.10\uparrow)$\\
    \bottomrule
    \end{tabular}
    
    \vspace{-1em}
    \label{tab:RM_eval}
\end{table*}

In this section, we evaluate the performance of Surfer in progressive instruction reasoning tasks. In the SeaWave benchmark, the training and testing environments mainly include instructions and scenes (\textit{\textit{i.e.}}, objects, positions, rotation information, etc.). Because the scenarios are regenerated each time, the scene is unseen for each level of task.
At the time of testing, since the level 1 instructions are in the form of “verb + noun”, they are seen. For levels 2, 3, and 4, human-like natural language instructions are used, so they are unseen, and the corresponding number of instructions are 80, 172, and 226, respectively.

The success rates of Surfer's manipulation on four progressive inference tasks are shown in Table \ref{tab:RM_eval}. As shown in Table \ref{tab:RM_eval}, compared to other baselines, Surfer has achieved a significant improvement in the success rate of manipulation on all tasks. Specifically, compared to RT-1, Surfer's manipulation success rate increased by an average of 7.1\% on four levels of progressive reasoning tasks. 
It is worth noting that Surfer still has a 3.15\% (37.89 \textit{vs.} 34.74) improvement in the most challenging level 4 tasks.

We have shown some examples of robot manipulation in Figure \ref{fig:quantative}. 
We compared the reasoning abilities of RT-1 and Surfer in terms of positional relationships, spatial size, and color appearance. The results showed that Surfer demonstrated excellent cross-modal reasoning and manipulation abilities.
Specifically, in the third task, if it is limited to green objects, RT-1 is also correct, but we require an object with a green cap, so only the object that Surfer grabs is correct in the end.

\vspace{-1.5em}
\subsection{Model Robustness}
\vspace{-0.5em}
\label{sec:generalization_results}
Here, we evaluate the robustness of Surfer in different scenarios using the average manipulation success rate of the model on level 2, 3, and 4 tasks.
New scenarios include unseen backgrounds (\textit{i.e.}, two unseen tables), changing light intensity, and more distractions (\textit{i.e.}, more objects and $4 \leq n\leq 6$). Examples of the new scenarios are depicted in the Appendix. 
We still adopt the instructions used in Section \ref{sec:results_rm}, but the scenarios have been changed.

As shown in Table \ref{tab:robustness}, we found that more distractors significantly reduced the success rate of model manipulation.
In contrast, Surfer is still significantly better than other baselines under all interference factors, and there is no obvious performance loss on the unseen background and changing lights, which shows that Surfer has good robustness with the help of the world model.

\begin{table}[t]
    \centering
    \caption{Performance comparison on seen scenarios, different backgrounds, changing lights, and more distractors (\%).}
    \vspace{-0.5em}
    \begin{tabular}{l|cccc}
    \toprule
        Model & Seen & \makecell{Unseen backgrounds} & \makecell{Changing lights}  & Distractors \\ \hline
        BC-Z \cite{pmlr-v164-jang22a} & 27.02 & 19.17 &  18.33 &  21.67\\
        Gato \cite{reed2022generalist} & 24.56 & 20.83 & 23.33 & 16.67 \\
        RT-1 \cite{rt12022arxiv} & 41.05 & 38.33 & 40.83  &  35.00 \\ \hline
        Surfer & $\mathbf{48.07}$  & $\mathbf{46.67}$ & $\mathbf{45.83}$ & $\mathbf{40.83}$\\
    \bottomrule
    \end{tabular}
    
    \label{tab:robustness}
\end{table}

\vspace{-1.5em}
\subsection{Ablation Studies}
\begin{figure}[t]
    \centering
    \subfloat[Visual comparison.]{\includegraphics[width=0.38 \linewidth]{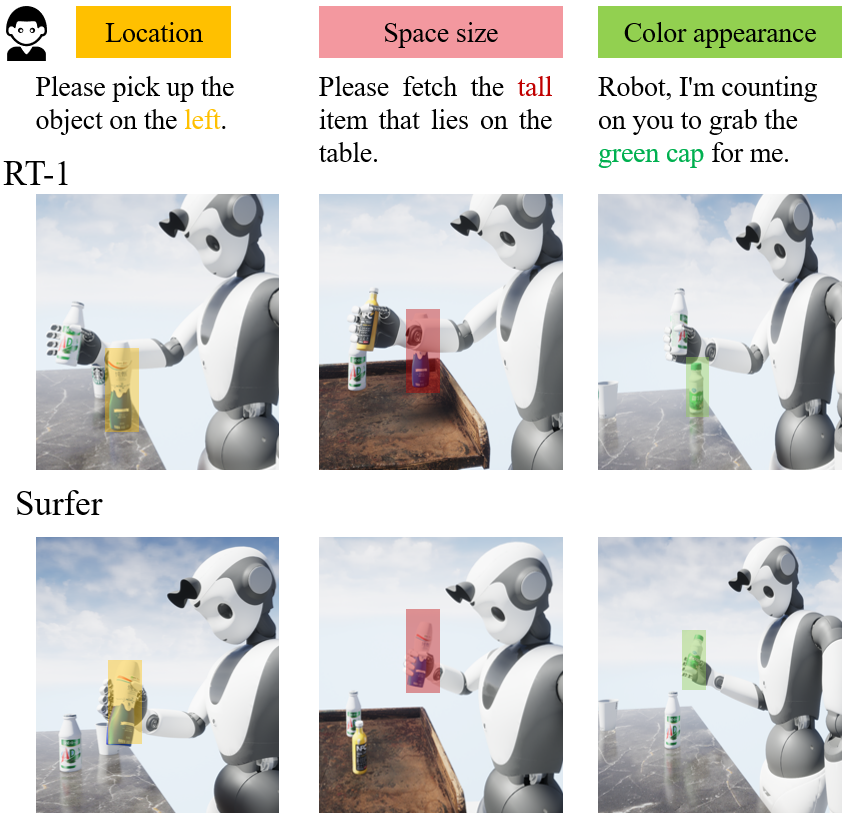}
    \label{fig:quantative}
    }
    \subfloat[Ablation results.]{\includegraphics[width=0.62 \linewidth]{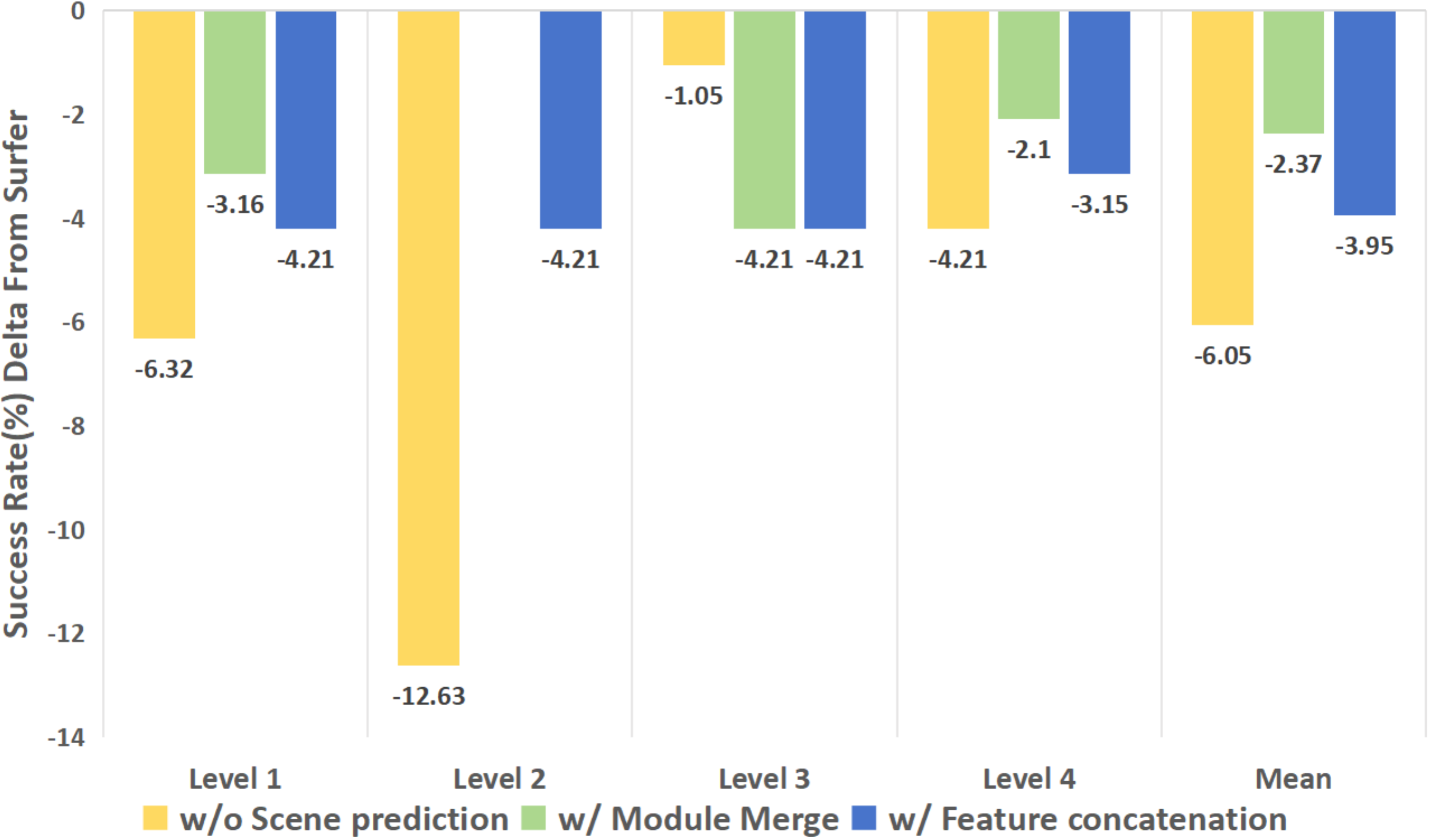}
    \label{fig:ablation}
    }
    \vspace{-0.5em}
    \caption{
    (a) Manipulation instances of RT-1 and Surfer on level 4 tasks. They evaluated the model's manipulation and reasoning abilities in terms of position, space, and appearance. Among them, the object with a background box is the target object of the current instruction. 
    (b) The ablation experiments of scene prediction, module merging, and feature concatenation.
    }
    
    \vspace{-2em}
\end{figure}

\vspace{-0.5em}
\label{sec:ablation studies}
In this section, we ablate the main modules of Surfer, including multi-modal feature fusion strategy and world model construction.
The corresponding ablation experimental results are shown in Figure \ref{fig:ablation}.

\noindent\textbf{Multi-Modal Fusion Strategy} 
In addition to using a transformer decoder for multi-modal feature fusion, we also attempted to directly concatenate features of different modalities (\textit{i.e.}, $\hat{I}_t, \hat{P}$ and $\hat{S}_t$) into the transformer layer for encoding.
Specifically, as shown in the blue bar graph in Figure \ref{fig:ablation}, compared with the original model, the concatenation feature reduces the success rate of the model's manipulation on the four-level tasks by an average of 3.95\%.
This may be because the concatenated feature fusion method greatly reduces the ability of image features to learn association information from instructions and robot state features, which is very unfavorable for the action prediction of instructions.

\noindent\textbf{World Model} 
Here, we perform ablation on the scene prediction module and prediction method in the world model.
Specifically, we first remove the scene prediction module.
As shown in the yellow bar graph in Figure \ref{fig:ablation}, removing the scene prediction module greatly weakens the model's ability to predict robot manipulation. 

Specifically, the model that removed the scene prediction module had an average decrease of 6.05\% in the success rate of manipulation on four levels of tasks.
This once again demonstrates the rationality of using the world model to predict the future scene and action of the robot and help the model learn manipulation consistent with world logical knowledge.

Secondly, we change the way of scene prediction, such as directly using the action prediction module to predict actions and scenes simultaneously.
As shown in the green bar graph in Figure \ref{fig:ablation},  to guide the model in predicting future scenarios is significantly better than using text instructions. That is, $I_{t-k:t}\xrightarrow{A_{t+1}} I_{t+1}$ makes more sense than $I_{t-k:t}\xrightarrow{P} I_{t+1}$. From a model perspective, this may mainly be because actions are closer to images than text instruction.


\section{Discussion}
\vspace{-0.5em}
\noindent\textbf{Conclusion}
Overall, in this work, we first constructed a realistic digital twin simulator for restaurant scenes based on UE5, which includes common objects in daily scenes and supports the Mujoco physics engine instead of relying on existing interfaces.
Then, based on this, we constructed a universal robot manipulation benchmark with progressive reasoning, SeaWave. It covers common robot manipulation tasks and constructs a standard evaluation platform for cross-modal robot manipulation.
Finally, we propose a new world model tailored for cross-modal robot manipulation, Surfer. Compared with other famous baselines, it maintains a leading manipulation success rate in all four levels of progressive reasoning tasks proposed.

\noindent\textbf{Limitation}
First, robots perform tasks using a single hand. Dual-handed operation is the direction of future exploration, which requires more precision in control over positioning and applied force. Moreover, while our simulator primarily interacts with rigid objects, the ability to manipulate soft materials is also important. For this reason, adding force sensors seems to be a useful option, and the introduction of more modal information will be helpful for decision-making.

\section{Acknowledgements}
This work was supported in part by National Key R\&D Program of China under Grant No. 2020AAA0109700, National Natural Science Foundation of China (NSFC) under Grant No.U19A2073 and No.61976233, Guangdong Province Basic and Applied Basic Research (Regional Joint Fund-Key) Grant No.2019B1515120039, National Natural Science Foundation of Guangdong Province (Grant No.2022A1515011835), China Postdoctoral Science Foundation funded project (Grant No. 2021M703687), Shenzhen Fundamental Research Program (Project No. RCYX20200714114642083, No. JCYJ20190807154211365), CAAI-Huawei MindSpore Open Fund. We also thank MindSpore for the partial support of this work, which is a new deep learning computing framework. 

\clearpage  

%
%
\bibliographystyle{splncs04}
\bibliography{main}

\begin{thebibliography}{10}
\providecommand{\url}[1]{\texttt{#1}}
\providecommand{\urlprefix}{URL }
\providecommand{\doi}[1]{https://doi.org/#1}

\bibitem{alayrac2022flamingo}
Alayrac, J.B., Donahue, J., Luc, P., Miech, A., Barr, I., Hasson, Y., Lenc, K., Mensch, A., Millican, K., Reynolds, M., et~al.: Flamingo: a visual language model for few-shot learning. Advances in Neural Information Processing Systems  \textbf{35},  23716--23736 (2022)

\bibitem{assran2023selfsupervised}
Assran, M., Duval, Q., Misra, I., Bojanowski, P., Vincent, P., Rabbat, M., LeCun, Y., Ballas, N.: Self-supervised learning from images with a joint-embedding predictive architecture (2023)

\bibitem{rt12022arxiv}
Brohan, A., Brown, N., Carbajal, J., Chebotar, Y., Dabis, J., Finn, C., Gopalakrishnan, K., Hausman, K., Herzog, A., Hsu, J., et~al.: Rt-1: Robotics transformer for real-world control at scale. arXiv preprint arXiv:2212.06817  (2022)

\bibitem{9812302}
Chao, Y.W., Paxton, C., Xiang, Y., Yang, W., Sundaralingam, B., Chen, T., Murali, A., Cakmak, M., Fox, D.: Handoversim: A simulation framework and benchmark for human-to-robot object handovers. In: 2022 International Conference on Robotics and Automation (ICRA). pp. 6941--6947 (2022). \doi{10.1109/ICRA46639.2022.9812302}

\bibitem{koala_blogpost_2023}
Chiang, W.L., Li, Z., Lin, Z., Sheng, Y., Wu, Z., Zhang, H., Zheng, L., Zhuang, S., Zhuang, Y., Gonzalez, J.E., Stoica, I., Xing, E.P.: Vicuna: An open-source chatbot impressing gpt-4 with 90\%* chatgpt quality (March 2023), \url{https://lmsys.org/blog/2023-03-30-vicuna/}

\bibitem{chowdhery2022palm}
Chowdhery, A., Narang, S., Devlin, J., Bosma, M., Mishra, G., Roberts, A., Barham, P., Chung, H.W., Sutton, C., Gehrmann, S., et~al.: Palm: Scaling language modeling with pathways. arXiv preprint arXiv:2204.02311  (2022)

\bibitem{du2022glm}
Du, Z., Qian, Y., Liu, X., Ding, M., Qiu, J., Yang, Z., Tang, J.: Glm: General language model pretraining with autoregressive blank infilling. In: Proceedings of the 60th Annual Meeting of the Association for Computational Linguistics (Volume 1: Long Papers). pp. 320--335 (2022)

\bibitem{ebert2018visual}
Ebert, F., Finn, C., Dasari, S., Xie, A., Lee, A., Levine, S.: Visual foresight: Model-based deep reinforcement learning for vision-based robotic control. arXiv preprint arXiv:1812.00568  (2018)

\bibitem{gal2016improving}
Gal, Y., McAllister, R., Rasmussen, C.E.: Improving pilco with bayesian neural network dynamics models. In: Data-efficient machine learning workshop, ICML. vol.~4, p.~25 (2016)

\bibitem{gao2022dialfred}
Gao, X., Gao, Q., Gong, R., Lin, K., Thattai, G., Sukhatme, G.S.: Dialfred: Dialogue-enabled agents for embodied instruction following. IEEE Robotics and Automation Letters  \textbf{7}(4),  10049--10056 (2022)

\bibitem{gu2023maniskill2}
Gu, J., Xiang, F., Li, X., Ling, Z., Liu, X., Mu, T., Tang, Y., Tao, S., Wei, X., Yao, Y., Yuan, X., Xie, P., Huang, Z., Chen, R., Su, H.: Maniskill2: A unified benchmark for generalizable manipulation skills. In: International Conference on Learning Representations (2023)

\bibitem{ha2018recurrent}
Ha, D., Schmidhuber, J.: Recurrent world models facilitate policy evolution. Advances in neural information processing systems  \textbf{31} (2018)

\bibitem{hafner2019dream}
Hafner, D., Lillicrap, T., Ba, J., Norouzi, M.: Dream to control: Learning behaviors by latent imagination. arXiv preprint arXiv:1912.01603  (2019)

\bibitem{hafner2020mastering}
Hafner, D., Lillicrap, T., Norouzi, M., Ba, J.: Mastering atari with discrete world models. arXiv preprint arXiv:2010.02193  (2020)

\bibitem{hao2023reasoning}
Hao, S., Gu, Y., Ma, H., Hong, J.J., Wang, Z., Wang, D.Z., Hu, Z.: Reasoning with language model is planning with world model. arXiv preprint arXiv:2305.14992  (2023)

\bibitem{he2022maskedMAE}
He, K., Chen, X., Xie, S., Li, Y., Doll{\'a}r, P., Girshick, R.: Masked autoencoders are scalable vision learners. In: Proceedings of the IEEE/CVF conference on computer vision and pattern recognition. pp. 16000--16009 (2022)

\bibitem{9001253}
James, S., Ma, Z., Arrojo, D.R., Davison, A.J.: Rlbench: The robot learning benchmark \& learning environment. IEEE Robotics and Automation Letters  \textbf{5}(2),  3019--3026 (2020). \doi{10.1109/LRA.2020.2974707}

\bibitem{pmlr-v164-jang22a}
Jang, E., Irpan, A., Khansari, M., Kappler, D., Ebert, F., Lynch, C., Levine, S., Finn, C.: Bc-z: Zero-shot task generalization with robotic imitation learning. In: Proceedings of the 5th Conference on Robot Learning. Proceedings of Machine Learning Research, vol.~164, pp. 991--1002. PMLR (08--11 Nov 2022), \url{https://proceedings.mlr.press/v164/jang22a.html}

\bibitem{alignjia2021scaling}
Jia, C., Yang, Y., Xia, Y., Chen, Y.T., Parekh, Z., Pham, H., Le, Q., Sung, Y.H., Li, Z., Duerig, T.: Scaling up visual and vision-language representation learning with noisy text supervision. In: International conference on machine learning. pp. 4904--4916. PMLR (2021)

\bibitem{jiang2023vima}
Jiang, Y., Gupta, A., Zhang, Z., Wang, G., Dou, Y., Chen, Y., Fei-Fei, L., Anandkumar, A., Zhu, Y., Fan, L.: Vima: General robot manipulation with multimodal prompts. In: Fortieth International Conference on Machine Learning (2023)

\bibitem{kalashnikov2018scalable}
Kalashnikov, D., Irpan, A., Pastor, P., Ibarz, J., Herzog, A., Jang, E., Quillen, D., Holly, E., Kalakrishnan, M., Vanhoucke, V., et~al.: Scalable deep reinforcement learning for vision-based robotic manipulation. In: Conference on Robot Learning. pp. 651--673. PMLR (2018)

\bibitem{kalashnikov2021mt}
Kalashnikov, D., Varley, J., Chebotar, Y., Swanson, B., Jonschkowski, R., Finn, C., Levine, S., Hausman, K.: Mt-opt: Continuous multi-task robotic reinforcement learning at scale. arXiv preprint arXiv:2104.08212  (2021)

\bibitem{kim2020learning}
Kim, S.W., Zhou, Y., Philion, J., Torralba, A., Fidler, S.: Learning to simulate dynamic environments with gamegan. In: Proceedings of the IEEE/CVF Conference on Computer Vision and Pattern Recognition. pp. 1231--1240 (2020)

\bibitem{LaValle1998RapidlyexploringRT}
LaValle, S.M.: Rapidly-exploring random trees : a new tool for path planning. The annual research report  (1998), \url{https://api.semanticscholar.org/CorpusID:14744621}

\bibitem{li2021igibson}
Li, C., Xia, F., Mart{\'\i}n-Mart{\'\i}n, R., Lingelbach, M., Srivastava, S., Shen, B., Vainio, K., Gokmen, C., Dharan, G., et~al.: igibson 2.0: Object-centric simulation for robot learning of everyday household tasks. arXiv preprint arXiv:2108.03272  (2021)

\bibitem{li2023vision}
Li, X., Liu, M., Zhang, H., Yu, C., Xu, J., Wu, H., Cheang, C., Jing, Y., Zhang, W., Liu, H., et~al.: Vision-language foundation models as effective robot imitators. arXiv preprint arXiv:2311.01378  (2023)

\bibitem{mishra2021cross}
Mishra, S., Khashabi, D., Baral, C., Hajishirzi, H.: Cross-task generalization via natural language crowdsourcing instructions. arXiv preprint arXiv:2104.08773  (2021)

\bibitem{mu2021maniskill}
Mu, T., Ling, Z., Xiang, F., Yang, D.C., Li, X., Tao, S., Huang, Z., Jia, Z., Su, H.: Maniskill: Generalizable manipulation skill benchmark with large-scale demonstrations. In: Thirty-fifth Conference on Neural Information Processing Systems Datasets and Benchmarks Track (Round 2) (2021)

\bibitem{maaz2023videochatgpt}
Muhammad~Maaz, Hanoona~Rasheed, S.K., Khan, F.: Video-chatgpt. \url{https://github.com/hanoonaR/Video-ChatGPT} (2023)

\bibitem{openai2023gpt4}
OpenAI: Gpt-4 technical report (2023)

\bibitem{ChatGPT}
OpenAI: Introducing chatgpt. In: https://openai.com/blog/chatgpt (2023)

\bibitem{padmakumar2021teach}
Padmakumar, A., Thomason, J., Shrivastava, A., Lange, P., Narayan-Chen, A., Gella, S., Piramuthu, R., Tur, G., Hakkani-Tur, D.: Teach: Task-driven embodied agents that chat (2021)

\bibitem{perez2018film}
Perez, E., Strub, F., De~Vries, H., Dumoulin, V., Courville, A.: Film: Visual reasoning with a general conditioning layer. In: Proceedings of the AAAI conference on artificial intelligence. vol.~32 (2018)

\bibitem{DBLP:journals/corr/abs-2103-00020}
Radford, A., Kim, J.W., Hallacy, C., Ramesh, A., Goh, G., Agarwal, S., Sastry, G., Askell, A., Mishkin, P., Clark, J., Krueger, G., Sutskever, I.: Learning transferable visual models from natural language supervision. CoRR  \textbf{abs/2103.00020} (2021), \url{https://arxiv.org/abs/2103.00020}

\bibitem{clipradford2021learning}
Radford, A., Kim, J.W., Hallacy, C., Ramesh, A., Goh, G., Agarwal, S., Sastry, G., Askell, A., Mishkin, P., Clark, J., et~al.: Learning transferable visual models from natural language supervision. In: International conference on machine learning. pp. 8748--8763. PMLR (2021)

\bibitem{DBLP:journals/corr/abs-1910-10683}
Raffel, C., Shazeer, N., Roberts, A., Lee, K., Narang, S., Matena, M., Zhou, Y., Li, W., Liu, P.J.: Exploring the limits of transfer learning with a unified text-to-text transformer. CoRR  \textbf{abs/1910.10683} (2019), \url{http://arxiv.org/abs/1910.10683}

\bibitem{2020t5}
Raffel, C., Shazeer, N., Roberts, A., Lee, K., Narang, S., Matena, M., Zhou, Y., Li, W., Liu, P.J.: Exploring the limits of transfer learning with a unified text-to-text transformer. Journal of Machine Learning Research  \textbf{21}(140),  1--67 (2020), \url{http://jmlr.org/papers/v21/20-074.html}

\bibitem{reed2022generalist}
Reed, S., Zolna, K., Parisotto, E., Colmenarejo, S.G., Novikov, A., Barth-Maron, G., Gimenez, M., Sulsky, Y., Kay, J., Springenberg, J.T., et~al.: A generalist agent. arXiv preprint arXiv:2205.06175  (2022)

\bibitem{ryoo2021tokenlearner}
Ryoo, M., Piergiovanni, A., Arnab, A., Dehghani, M., Angelova, A.: Tokenlearner: Adaptive space-time tokenization for videos. Advances in Neural Information Processing Systems  \textbf{34},  12786--12797 (2021)

\bibitem{sekar2020planning}
Sekar, R., Rybkin, O., Daniilidis, K., Abbeel, P., Hafner, D., Pathak, D.: Planning to explore via self-supervised world models. In: International Conference on Machine Learning. pp. 8583--8592. PMLR (2020)

\bibitem{seo2023masked}
Seo, Y., Hafner, D., Liu, H., Liu, F., James, S., Lee, K., Abbeel, P.: Masked world models for visual control. In: Conference on Robot Learning. pp. 1332--1344. PMLR (2023)

\bibitem{9636667}
Shen, B., Xia, F., Li, C., Martín-Martín, R., Fan, L., Wang, G., Pérez-D’Arpino, C., et~al.: igibson 1.0: A simulation environment for interactive tasks in large realistic scenes. In: 2021 IEEE/RSJ International Conference on Intelligent Robots and Systems (IROS). pp. 7520--7527 (2021). \doi{10.1109/IROS51168.2021.9636667}

\bibitem{ALFRED20}
Shridhar, M., Thomason, J., Gordon, D., Bisk, Y., Han, W., Mottaghi, R., Zettlemoyer, L., Fox, D.: {ALFRED: A Benchmark for Interpreting Grounded Instructions for Everyday Tasks}. In: The IEEE Conference on Computer Vision and Pattern Recognition (CVPR) (2020), \url{https://arxiv.org/abs/1912.01734}

\bibitem{ALFWorld20}
Shridhar, M., Yuan, X., C\^ot\'e, M.A., Bisk, Y., Trischler, A., Hausknecht, M.: {ALFWorld: Aligning Text and Embodied Environments for Interactive Learning}. In: Proceedings of the International Conference on Learning Representations (ICLR) (2021), \url{https://arxiv.org/abs/2010.03768}

\bibitem{pmlr-v164-srivastava22a}
Srivastava, S., Li, C., Lingelbach, M., Mart\'in-Mart\'in, R., Xia, F., Vainio, K.E., Lian, Z., Gokmen, C., Buch, S., Liu, K., et~al.: Behavior: Benchmark for everyday household activities in virtual, interactive, and ecological environments. In: Proceedings of the 5th Conference on Robot Learning. Proceedings of Machine Learning Research, vol.~164, pp. 477--490. PMLR (08--11 Nov 2022)

\bibitem{NEURIPS2021_021bbc7e}
Szot, A., Clegg, A., Undersander, E., Wijmans, E., Zhao, Y., Turner, J., Maestre, N., Mukadam, M., Chaplot, D.S., Maksymets, O., et~al.: Habitat 2.0: Training home assistants to rearrange their habitat. In: Advances in Neural Information Processing Systems. vol.~34, pp. 251--266. Curran Associates, Inc. (2021)

\bibitem{alpaca}
Taori, R., Gulrajani, I., Zhang, T., Dubois, Y., Li, X., Guestrin, C., Liang, P., Hashimoto, T.B.: Stanford alpaca: An instruction-following llama model. \url{https://github.com/tatsu-lab/stanford_alpaca} (2023)

\bibitem{touvron2023llama}
Touvron, H., Lavril, T., Izacard, G., Martinet, X., Lachaux, M.A., Lacroix, T., Rozière, B., Goyal, N., Hambro, E., Azhar, F., Rodriguez, A., Joulin, A., Grave, E., Lample, G.: Llama: Open and efficient foundation language models (2023)

\bibitem{vaswani2017attention}
Vaswani, A., Shazeer, N., Parmar, N., Uszkoreit, J., Jones, L., Gomez, A.N., Kaiser, {\L}., Polosukhin, I.: Attention is all you need. Advances in neural information processing systems  \textbf{30} (2017)

\bibitem{vuong2023openrt-x}
Vuong, Q., Levine, S., Walke, H.R., Pertsch, K., Singh, A., Doshi, R., Xu, C., Luo, J., Tan, L., Shah, D., et~al.: Open x-embodiment: Robotic learning datasets and rt-x models. In: Towards Generalist Robots: Learning Paradigms for Scalable Skill Acquisition@ CoRL2023 (2023)

\bibitem{NEURIPS2022_4eb33c53}
Wan, Y., Mao, J., Tenenbaum, J.: Handmethat: Human-robot communication in physical and social environments. In: Advances in Neural Information Processing Systems. vol.~35, pp. 12014--12026. Curran Associates, Inc. (2022), \url{https://proceedings.neurips.cc/paper_files/paper/2022/file/4eb33c53ed5b14ce9028309431f565cc-Paper-Datasets_and_Benchmarks.pdf}

\bibitem{wang2023drivedreamer}
Wang, X., Zhu, Z., Huang, G., Chen, X., Lu, J.: Drivedreamer: Towards real-world-driven world models for autonomous driving. arXiv preprint arXiv:2309.09777  (2023)

\bibitem{wang2023selfconsistency}
Wang, X., Wei, J., Schuurmans, D., Le, Q., Chi, E., Narang, S., Chowdhery, A., Zhou, D.: Self-consistency improves chain of thought reasoning in language models (2023)

\bibitem{wang2022self}
Wang, Y., Kordi, Y., Mishra, S., Liu, A., Smith, N.A., Khashabi, D., Hajishirzi, H.: Self-instruct: Aligning language model with self generated instructions. arXiv preprint arXiv:2212.10560  (2022)

\bibitem{wei2021finetuned}
Wei, J., Bosma, M., Zhao, V.Y., Guu, K., Yu, A.W., Lester, B., Du, N., Dai, A.M., Le, Q.V.: Finetuned language models are zero-shot learners. arXiv preprint arXiv:2109.01652  (2021)

\bibitem{wei2023chainofthought}
Wei, J., Wang, X., Schuurmans, D., Bosma, M., Ichter, B., Xia, F., Chi, E., Le, Q., Zhou, D.: Chain-of-thought prompting elicits reasoning in large language models (2023)

\bibitem{weller2020learning}
Weller, O., Lourie, N., Gardner, M., Peters, M.E.: Learning from task descriptions. arXiv preprint arXiv:2011.08115  (2020)

\bibitem{wu2023visual}
Wu, C., Yin, S., Qi, W., Wang, X., Tang, Z., Duan, N.: Visual chatgpt: Talking, drawing and editing with visual foundation models (2023)

\bibitem{wu2023unleashing}
Wu, H., Jing, Y., Cheang, C., Chen, G., Xu, J., Li, X., Liu, M., Li, H., Kong, T.: Unleashing large-scale video generative pre-training for visual robot manipulation. arXiv preprint arXiv:2312.13139  (2023)

\bibitem{wu2023daydreamer}
Wu, P., Escontrela, A., Hafner, D., Abbeel, P., Goldberg, K.: Daydreamer: World models for physical robot learning. In: Conference on Robot Learning. pp. 2226--2240. PMLR (2023)

\bibitem{xiang2020sapien}
Xiang, F., Qin, Y., Mo, K., Xia, Y., Zhu, H., Liu, F., Liu, M., Jiang, H., Yuan, Y., Wang, H., Yi, L., Chang, A.X., Guibas, L.J., Su, H.: Sapien: A simulated part-based interactive environment (2020)

\bibitem{yao2023tree}
Yao, S., Yu, D., Zhao, J., Shafran, I., Griffiths, T.L., Cao, Y., Narasimhan, K.: Tree of thoughts: Deliberate problem solving with large language models (2023)

\bibitem{yenamandra2023homerobot}
Yenamandra, S., Ramachandran, A., Yadav, K., Wang, A., Khanna, M., Gervet, T., Yang, T.Y., Jain, V., Clegg, A.W., Turner, J., Kira, Z., Savva, M., Chang, A., Chaplot, D.S., Batra, D., Mottaghi, R., Bisk, Y., Paxton, C.: Homerobot: Open-vocabulary mobile manipulation (2023)

\bibitem{zeng2020transporter}
Zeng, A., Florence, P., Tompson, J., Welker, S., Chien, J., Attarian, M., Armstrong, T., Krasin, I., Duong, D., Sindhwani, V., Lee, J.: Transporter networks: Rearranging the visual world for robotic manipulation. Conference on Robot Learning (CoRL)  (2020)

\bibitem{zhang2018deep}
Zhang, T., McCarthy, Z., Jow, O., Lee, D., Chen, X., Goldberg, K., Abbeel, P.: Deep imitation learning for complex manipulation tasks from virtual reality teleoperation. In: 2018 IEEE International Conference on Robotics and Automation (ICRA). pp. 5628--5635. IEEE (2018)

\bibitem{zheng2022vlmbench}
Zheng, K., Chen, X., Jenkins, O.C., Wang, X.: Vlmbench: A compositional benchmark for vision-and-language manipulation. Advances in Neural Information Processing Systems  \textbf{35},  665--678 (2022)

\bibitem{zhu2023minigpt}
Zhu, D., Chen, J., Shen, X., Li, X., Elhoseiny, M.: Minigpt-4: Enhancing vision-language understanding with advanced large language models. arXiv preprint arXiv:2304.10592  (2023)

\bibitem{robosuite2020}
Zhu, Y., Wong, J., Mandlekar, A., Mart\'{i}n-Mart\'{i}n, R., Joshi, A., Nasiriany, S., Zhu, Y.: robosuite: A modular simulation framework and benchmark for robot learning. In: arXiv preprint arXiv:2009.12293 (2020)

\bibitem{zitkovich2023rt2}
Zitkovich, B., Yu, T., Xu, S., Xu, P., Xiao, T., Xia, F., Wu, J., Wohlhart, P., Welker, S., Wahid, A., et~al.: Rt-2: Vision-language-action models transfer web knowledge to robotic control. In: 7th Annual Conference on Robot Learning (2023)

\end{thebibliography}
\end{document}